\pdfoutput=1

\documentclass[11pt]{article}

\usepackage{acl}

\usepackage{times}
\usepackage{latexsym}

\usepackage[T1]{fontenc}

\usepackage[utf8]{inputenc}

\usepackage{microtype}

\usepackage{inconsolata}

\usepackage{graphicx}
\usepackage[ruled,linesnumbered]{algorithm2e}
\usepackage{amsmath}
\usepackage{amssymb}
\usepackage{makecell}
\usepackage{multirow}
\usepackage{ulem}
\usepackage{colortbl}
\usepackage{booktabs}
\usepackage{soul, color, xcolor}
\usepackage{subfig}
\usepackage{pifont}
\usepackage{float}

\newcommand{\cmark}{\textcolor{teal}{\ding{51}}}
\newcommand{\xmark}{\textcolor{red}{\ding{55}}}
\title{Curriculum-style Data Augmentation for LLM-based Metaphor Detection}
  
 \author{Kaidi Jia\textsuperscript{$1$}\thanks{Equal contribution}, Yanxia Wu\textsuperscript{$1*$}, Ming Liu\textsuperscript{$2$} \and Rongsheng Li\textsuperscript{$1*$}\thanks{Corresponding author} \\
	\textsuperscript{$1$}College of Computer Science and Technology, Harbin Engineering University, Harbin 150001, China   \\
	\textsuperscript{$2$}Deakin University, Australia \\
	\texttt{dasheng@hrbeu.edu.cn}}

\begin{document}
\maketitle
\begin{abstract}
Recently, utilizing large language models (LLMs) for metaphor detection has achieved promising results. However, these methods heavily rely on the capabilities of closed-source LLMs, which come with relatively high inference costs and latency. To address this, we propose a method for metaphor detection by fine-tuning open-source LLMs, effectively reducing inference costs and latency with a single inference step. Furthermore, metaphor detection suffers from a severe data scarcity problem, which hinders effective fine-tuning of LLMs. To tackle this, we introduce \textbf{C}urriculum-style \textbf{D}ata \textbf{A}ugmentation (\textbf{CDA}). Specifically, before fine-tuning, we evaluate the training data to identify correctly predicted instances for fine-tuning, while incorrectly predicted instances are used as seed data for data augmentation. This approach enables the model to quickly learn simpler knowledge and progressively acquire more complex knowledge, thereby improving performance incrementally. Experimental results demonstrate that our method achieves state-of-the-art performance across all baselines. Additionally, we provide detailed ablation studies to validate the effectiveness of CDA.
\end{abstract}

\section{Introduction}

As large language models (LLMs) continue to grow in capability, they have been widely applied across various domains of natural language processing (NLP). These include tasks such as instruction following  \citep{brown2020language, ouyang2024training, wang-etal-2023-self-instruct}, code generation  \citep{chen2021evaluating, nijkamp2023codegen, luo2023wizardcoder}, and mathematical reasoning \citep{cobbe2021training, zhou2023least-to-most, imani-etal-2023-mathprompter}, where they have demonstrated outstanding performance. Recent studies  \citep{tian-etal-2024-theory, yang-etal-2024-chatgpts} have explored leveraging LLMs for metaphor detection task, employing various prompt designs to maximize the performance of proprietary LLMs and achieving remarkable results across multiple metaphor detection datasets. However, these methods often require multiple inference steps, leading to higher computational costs and increased inference latency. Therefore, we focus on developing a supervised method to enhance the performance of open-source LLMs on metaphor detection task, aiming to reduce both inference costs and latency by relying on a single inference step.

Research on metaphor detection using fine-tuned LLMs is still in its early stages, with the primary challenge stemming from the severe data scarcity problem inherent in metaphor detection task. When training with BERT, previous studies \citep{zhou-etal-2023-clcl, jia-li-2024-metaphor} have introduced curriculum learning to mitigate the impact of data scarcity. However, while this method enables more effective utilization of limited data, it only alleviates the problem of data scarcity.

To address the data scarcity problem in metaphor detection and enable more effective fine-tuning of LLMs, we designed a dedicated data augmentation template for metaphor detection and employed an iterative approach to generate useful data. This process includes two key components: (1) \textbf{Data Generation:} For each sentence, we use three different prompts to generate diverse data from various perspectives.  (2) \textbf{Useful Data Selection:} Among the generated data, we evaluate all samples using the model to be trained, selecting correctly predicted data for fine-tuning while using incorrectly predicted data as seed data for subsequent data generation.

This method incorporates the concept of curriculum learning  \citep{bengio2009curriculum}. Initially, the model learns from simpler data, and as its capability improves, it progressively tackles more challenging, previously unlearned data. Consequently, we refer to this method as \textbf{C}urriculum-style \textbf{D}ata \textbf{A}ugmentation (\textbf{CDA}). By employing CDA, the model can first acquire foundational knowledge when its capacity is relatively weak, leading to rapid performance improvements. As the model's ability grows, it can gradually learn more complex knowledge, resulting in continuous performance enhancement. Our main contributions are as follows: 

\begin{itemize}
	\item[$\bullet$]To the best of our knowledge, we are the first to enhance the performance of open-source LLMs on metaphor detection task through fine-tuning. Additionally, we are the first to introduce data augmentation into metaphor detection task. By performing inference in a single step, we significantly reduce the cost and latency of inference.
	\item[$\bullet$]We propose a novel data augmentation method. Combining the concept of curriculum learning, we propose \textbf{C}urriculum-style \textbf{D}ata \textbf{A}ugmentation (\textbf{CDA}), an iterative method to increasingly generate useful data. Our method significantly addresses the data scarcity problem in metaphor detection task, enabling more effective model fine-tuning.
	\item[$\bullet$]Experimental results demonstrate that our method outperforms all approaches based on both BERT-based methods and LLM-based methods, achieving state-of-the-art performance.  Compared to other LLM-based approaches, our method requires the fewest model calls during inference, highlighting the efficiency of fine-tuning. Furthermore, by iteratively applying data augmentation, our model continuously improves its performance on metaphor detection task, showcasing the effectiveness of CDA.
\end{itemize}

\begin{figure*}
	\centering
	\includegraphics[width=\linewidth]{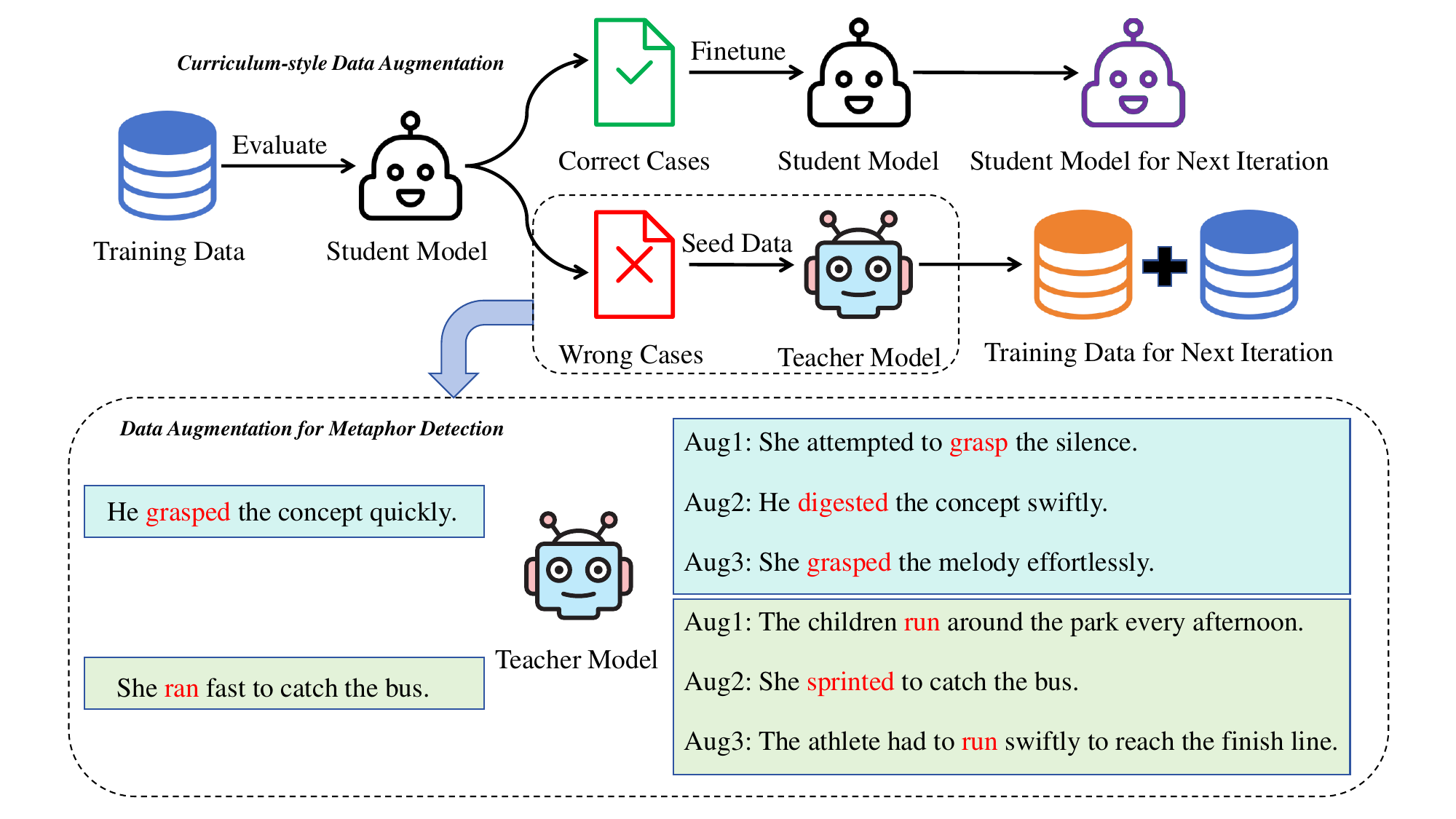}
	\caption{Structures of Our Method. The top half shows the process of Curriculum-style Data Augmentation (CDA), and the bottom half shows examples of data augmentation, metaphorical examples in blue boxes and non-metaphorical examples in green boxes.}
	\label{fig1}
\end{figure*}

\section{Related Works}

\subsection{Metaphor Detection}

Metaphors can make expressions more vivid and evocative, which is why they often appear in various literary works \citep{Lakoff1980}. Despite the challenge of identifying metaphors due to their lack of fixed structure, they are closely related to multiple NLP tasks, such as machine translation \citep{Shi2014TranslationAA, mao-etal-2018-word} and sentiment analysis \citep{socher-etal-2013-recursive, dankers-etal-2019-modelling}. Therefore, accurately identifying metaphors is of significant importance in the field of NLP.

Early works primarily focused on feature extraction and analysis from corpora \citep{shutova-etal-2010-metaphor, turney-etal-2011-literal, shutova-sun-2013-unsupervised, broadwell2013using, tsvetkov-etal-2014-metaphor} or word vectors \citep{shutova-etal-2016-black, bulat-etal-2017-modelling, rei-etal-2017-grasping}. However, due to the limitations in corpus size or vocabulary, these methods did not achieve excellent results. Later work \citep{gao-etal-2018-neural, wu-etal-2018-neural} employed RNNs to extract contextual information, improving performance by leveraging contextual clues. However, due to the inherent limitations of RNNs in capturing complex context, these methods struggled to identify intricate metaphors. With the advent of Transformer models \citep{vaswani2017attention}, pre-trained models like BERT \citep{devlin-etal-2019-bert} and RoBERTa \citep{liu2019roberta} have been able to better utilize contextual information, gradually becoming mainstream methods for metaphor detection \citep{gong-etal-2020-illinimet, su-etal-2020-deepmet, choi-etal-2021-melbert, zhang-liu-2022-metaphor, zhang-liu-2023-adversarial, zhou-etal-2023-clcl, jia-li-2024-metaphor}.

Recent approaches have attempted to introduce LLMs into the metaphor detection task. \citet{tian-etal-2024-theory} built a knowledge graph based on metaphor theories to guide LLMs in metaphor detection. \citet{yang-etal-2024-chatgpts} improved metaphor detection by having LLMs generate common verb collocations, thereby better utilizing metaphor theory. Unlike these prompt-based methods, we enhance performance in metaphor detection by fine-tuning open-source LLMs. Compared to other methods, our approach requires only a single call to the LLMs for metaphor detection, significantly reducing the cost and latency of inference.

\subsection{Data Augmentation for LLM}\label{otheraug}

Data augmentation is a common method to address the problem of data scarcity and has been widely applied in tasks such as instruction following \citep{meng2022generating, wang-etal-2023-self-instruct}, question answering \citep{ding-etal-2023-enhancing, chen-etal-2024-minprompt}, and mathematical reasoning \citep{yu2024metamath, li-etal-2024-mugglemath}, achieving good results. Given that metaphor detection also suffers from severe data scarcity, we introduce data augmentation into the metaphor detection task.

In this paper, we combine data augmentation with the concept of curriculum learning and propose \textbf{C}urriculum-style \textbf{D}ata \textbf{A}ugmentation (\textbf{CDA}). Specifically, after performing data augmentation, we use the model to be trained to evaluate the generated data. The correctly predicted data is selected for fine-tuning, while the incorrectly predicted data is used as seed data for generating the data for next iteration. The advantage of this method is that it allows for the selection of data that is beneficial for training, avoiding the waste of computational resources on simpler data. Moreover, the model can learn the simpler knowledge in the data during the early stages of training, rapidly improving performance, and as performance improves, it gradually learns more difficult data, continuously enhancing its performance.

\section{Problem Definition}

Given a sentence [Sent] and a target word [Tar] within it, our task is to determine whether the target word is used metaphorically in the sentence. To facilitate fine-tuning and prediction with large language models (LLMs) for this task, we reorganize the original data into the question-answer pair format shown in Table \ref{form}. In this format, if the target word in the sentence carries a metaphorical meaning, the output is "Yes"; otherwise, the output is "No."

\begin{table}[!t]\footnotesize
\centering
\small

\begin{tabular}{|p{0.95\linewidth}|}
\hline
\rule{-2pt}{12pt}
\textbf{Instruction:}\\
Is the word '[Tar]'  in the sentence '[Sent]' used metaphorically?     Please answer with 'Yes' or 'No' only.\\
\textbf{Input:} Null\\
\textbf{Output:} Yes/No\\

\hline
\end{tabular}

\caption{Reorganize the original data into question-answer pairs.}
\label{form}
\vspace{-5mm}
\end{table}

\section{Method}

We aim to address the data scarcity problem in the metaphor detection task through data augmentation, in order to better fine-tune open-source LLMs. However, some data may be overly simple or contain significant noise, and augmenting such data does not improve the performance of LLMs \citep{chen2024alpagasus}. To tackle this issue, we combine the concept of curriculum learning and propose \textbf{C}urriculum-style \textbf{D}ata \textbf{A}ugmentation  (\textbf{CDA}). As shown in Figure \ref{fig1}, Our method consists of two parts: (1) Data Augmentation for Metaphor Detection; (2) Curriculum-style Data Augmentation.

\subsection{Data Augmentation for Metaphor Detection}\label{dataaug}

For each sentence, we use three different methods for data augmentation. Specifically, we design three different prompts for metaphorical and literal data to ensure the diversity of generated data. For \textbf{metaphorical data}, we have: 1) directly generate sentences that include the metaphorical usage of the target word; 2) retain the context, replace the target word, and ensure that the replaced word still uses a metaphorical meaning; 3) retain the metaphorical usage of the target word and replace the context. For \textbf{literal data}, we have: 1) directly generate sentences that include the literal usage of the target word; 2) retain the context, replace the target word, and ensure that the replaced word still uses a literal meaning; 3) retain the literal usage of the target word and replace the context. The detailed prompts can be found in the appendix \ref{prompts}.

To ensure that the augmented data is of high quality, we introduce a powerful LLM as the teacher model $M_T$, which is responsible for generating data for the student model $M_S$ to be trained. Since the metaphor detection task is quite challenging, we need to ensure that the teacher model $M_T$ has a good understanding of the data to be augmented; otherwise, significant noise may appear in the generated data. 

Initially, we use the teacher model $M_T$ to evaluate all the data and select the correctly predicted data as the initial data, while the incorrectly predicted data is discarded. This is because the incorrectly predicted data can be categorized into two types: 1) data with significant noise, and 2) data that is too difficult. The former leads to noise in the data generated by the teacher model $M_T$, and the latter indicates that the teacher model $M_T$ lacks sufficient understanding of the data, which may result in incorrect data generation. Therefore, we discard all incorrectly predicted data and retain only the correctly predicted data as the initial data.

\begin{algorithm}
	\caption{Curriculum-style Data Augmentation}\label{algorithm}
	\KwIn{Dataset $\mathbb{D}^0$, Teacher Model $M_T$, Student Model $M_S^0$}
	\KwOut{Fine-tuned Student Model $M_S^N$}
	\For{n=1; n<=N}{
		$\mathbb{D}_{correct}^{n-1}, \mathbb{D}_{wrong}^{n-1} \leftarrow Evaluate(M_S^{n-1}, \mathbb{D}^{n-1})$\;
		$M_S^n \leftarrow Finetune(M_S^{n-1}, \mathbb{D}_{correct}^{n-1})$\;
		$Aug^n \leftarrow Generate(M_T, \mathbb{D}_{wrong}^{n-1})$\;
		$\mathbb{D}^n \leftarrow \mathbb{D}^{n-1} + Aug^n$\;
		$n \leftarrow n+1$\;
	}
	\Return{$M_S^N$}
\end{algorithm}

\subsection{Curriculum-style Data Augmentation}

We combine data augmentation with the concept of curriculum learning and propose \textbf{C}urriculum-style \textbf{D}ata \textbf{A}ugmentation (\textbf{CDA}). By iteratively learning relatively simple data and augmenting more difficult data, the student model $M_S$ can quickly learn simpler knowledge and gradually acquire more difficult knowledge. The algorithm for CDA is shown in Algorithm \ref{algorithm}.

Specifically, before each iteration of fine-tuning, we use the student model $M_S$ to evaluate the data for that iteration. For the data predicted correctly by the model, we consider the model to have a good understanding of these data and believe it can quickly learn useful knowledge from them. Therefore, we use these data for the current iteration of fine-tuning. For the data predicted incorrectly by the model, we assume that the model still does not fully understand these data and will face significant difficulty in learning them. Thus, we treat them as seed data for the data augmentation. After each iteration of data augmentation, we remove duplicate data from all generated data, and then combine the augmented data with the original data to form the training data for the next iteration. With this approach, we enable the model to quickly learn knowledge from relatively simple data, and as the model's capability improves, gradually learn more difficult data, continuously enhancing its performance.

\section{Experiments}

\subsection{Datasets}

\begin{table}
	\centering
	\resizebox{\linewidth}{!}{
		\begin{tabular}{ccccc}
			\Xhline{1.2pt}
			\rule{0pt}{12pt}
			\textbf{Dataset} & \textbf{\#Instance} & \textbf{\%M} & \textbf{\#Len} & \textbf{\#Samp}\\
			\hline
			\rule{0pt}{12pt}
			VUA Verb & 13,608 & 23.89 & 20.2 & 6000\\
			\hline
			\rule{0pt}{12pt}
			MOH-X & 647 & 48.69 & 8.0 & 300\\
			\rule{0pt}{12pt}
			TroFi & 3,737 & 43.54 & 28.3 & 300\\
			\Xhline{1.2pt}
	\end{tabular}}
	\caption{\label{dataset}
		 Statistics of datasets. \textbf{\#Instance}: Number of instances. \textbf{\%M}: Percentage of metaphorical instances. \textbf{\#Len}: Average length of instances. \textbf{\#Samp}: Number of sampled instances for training or testing. Among them, the VUA Verb statistics are the results after filtering the training set, and the filtering method is shown in \ref{dataaug}.}
\end{table}

\begin{table*}[t]
	\centering

		\begin{tabular}{lccccc}
			\Xhline{1.2pt}
			\rule{0pt}{15pt}
			\multirow{2}{*}{\textbf{Method}}
			& \multirow{2}{*}{\textbf{\#Call}} & \multicolumn{2}{c}{\textbf{MOH-X}} & \multicolumn{2}{c}{\textbf{TroFi}}\\ 
			\cline{3-6}
			\rule{0pt}{10pt}
			& & Acc. & F1 & Acc. & F1 \\
			\hline
			\rowcolor{gray!15}
			\multicolumn{6}{l}{\textit{BERT-based methods}}\\
			MelBERT \citep{choi-etal-2021-melbert} & 1 & 77.88 $\pm$ 0.83 & 77.89 $\pm$ 0.83 & 62.36 $\pm$ 1.51 & 62.89 $\pm$ 1.29 \\
			MisNet \citep{zhang-liu-2022-metaphor} & 1 & 77.08 $\pm$ 1.12 & 77.11 $\pm$ 1.13 & 62.01 $\pm$ 0.64 & 62.67 $\pm$ 0.54 \\
			AdMul \citep{zhang-liu-2023-adversarial} & 1 & 79.74 $\pm$ 0.44 & 79.89 $\pm$ 0.42 & 60.54 $\pm$ 1.43 & 62.67 $\pm$ 0.98 \\
			MiceCL \citep{jia-li-2024-metaphor} & 1 & 79.33 $\pm$ 1.70 & 79.06 $\pm$ 0.72 & 63.11 $\pm$ 1.78 & \uline{\textit{67.58 $\pm$ 1.12}} \\
			\rowcolor{gray!15}
			\multicolumn{6}{l}{\textit{LLM-based methods}}\\
			BCTK \citep{yang-etal-2024-chatgpts} & 1 & 53.58 $\pm$ 0.72 & 67.70 $\pm$ 0.34 & 48.48 $\pm$ 0.43 & 64.03 $\pm$ 0.23 \\
			TSI \citep{tian-etal-2024-theory} & 3 & \uline{\textit{82.59 $\pm$ 2.22}} & \uline{\textit{82.93 $\pm$ 1.94}} & \uline{\textit{66.07 $\pm$ 1.11}} & 66.89 $\pm$ 1.13 \\
			\rowcolor{gray!15}
			\multicolumn{6}{l}{\textit{Our methods}}\\
			Llama 3.1 8B Instruct & 1 & 70.00 $\pm$ 1.41 & 67.07 $\pm$ 1.59 & 60.44 $\pm$ 1.44 & 66.54 $\pm$ 1.25 \\
			Llama 3.1 8B Instruct (CDA) & 1 & \textbf{86.11 $\pm$ 0.27} & \textbf{86.57 $\pm$ 0.30} & \textbf{69.11 $\pm$ 0.27} & \textbf{73.62 $\pm$ 0.29}\\
			\Xhline{1.2pt} 
	\end{tabular}
	
	\caption{\label{mainresults}
		Results on MOH-X and TroFi. Best in bold and second best in italic underlined. \textbf{\#Call} represents the number of times the model needs to be called to test a single instance.}
\end{table*}

We fine-tune the model on the train dataset using CDA and perform zero-shot testing on the test dataset. The statistical information of the dataset is shown in Table \ref{dataset}. The statistical information of the dataset after augmented can be found in the appendix \ref{statistic}.

\subsubsection{Train Datasets}

\noindent\textbf{VUA Verb} \citep{Steen2010AMF}: The dataset is a verb subset of the largest metaphor detection dataset, VUA ALL \citep{Steen2010AMF}, collected by VUA from the BNCBaby corpus. The initial VUA Verb dataset comprised a total of 3,251 metaphorical instances and 10,357 non-metaphorical instances. To ensure a balanced distribution of the initial training data, we randomly selected 3,000 metaphorical instances and 3,000 non-metaphorical instances. The random selection was performed using a fixed seed of 42 to ensure reproducibility.

\subsubsection{Test Datasets}

\noindent\textbf{MOH-X} \citep{mohammad-etal-2016-metaphor}: The dataset is collected from WordNet and contains 647 instances.

\noindent\textbf{TroFi} \citep{birke-sarkar-2006-clustering}: The dataset is collected from the Wall Street Journal and contains 3,737 instances.

For comparison with methods based on LLMs, we directly used the data selected from TSI \citep{tian-etal-2024-theory} without any modification, with 150 metaphorical instances and 150 non-metaphorical instances extracted from each dataset.

While our experiments were conducted only on the verb datasets, this does not imply that our method is exclusively applicable to verb metaphor detection. We extended our experimental scope to other parts of speech within the VUA ALL dataset, encompassing adjectives, adverbs, and nouns, which can be found in appendix \ref{vuaall}. 

\subsection{Baselines}

We adopt the state-of-the-art (SOTA) methods prior to the metaphor detection task as baselines, including BERT-based mothods and LLM-based mothods, as follows:

\textbf{MelBERT} \citep{choi-etal-2021-melbert}: This method designs a metaphor detection framework that fully utilizes SPV and MIP metaphor theories for metaphor detection.
\textbf{MisNet} \citep{zhang-liu-2022-metaphor}: The framework of this method is similar to that of MelBERT, but it expresses the literal meaning of the target word more precisely.
\textbf{AdMul} \citep{zhang-liu-2023-adversarial}: This method proposes a multi-task learning framework that makes fuller use of the MIP metaphor theories.
\textbf{MiceCL} \citep{jia-li-2024-metaphor}: This method uses curriculum learning to alleviate the data scarcity problem and further improves the metaphor detection framework of MelBERT.
\textbf{BCTK}\footnote{Since the author did not give the method a name, we use the acronym of the title to represent the method.} \citep{yang-etal-2024-chatgpts}: This method utilizes LLMs to generate common verb collocations to better leverage metaphor theory for metaphor detection.
\textbf{TSI} \citep{tian-etal-2024-theory}: This method constructs a knowledge graph based on metaphor theories to guide LLMs in metaphor detection.

\subsection{Implementation Details}

We use GPT-4o (gpt-4o-2024-08-06) \citep{gpt4} as the teacher model and Llama 3.1 8B Instruct \citep{llama3} as the student model. We perform three iterations of fine-tuning using CDA on the VUA Verb dataset, training for 3 epochs during each iteration. We use LoRA \citep{lora} with rank 8, alpha 16, and a starting learning rate 2e-4. After fine-tuning, we conduct zero-shot testing on the MOH-X and TroFi datasets. During testing, we perform three experiments on each dataset and report the average performance and standard deviation of the three results. All experiments are conducted on two NVIDIA RTX A6000 GPUs.

\subsection{Main Results}

Table \ref{mainresults} presents the comparison results between our method and the baselines. Compared to directly using Llama 3.1 8B Instruct for prediction, the fine-tuned results show a significant improvement. On both test datasets, our fine-tuned model achieved the best performance, whether compared to BERT-based methods or LLM-based methods. It is worth mentioning that the BERT-based methods were trained on the VUA ALL dataset, while we fine-tuned only on the verb subset of VUA ALL and achieved better results. This improvement is mainly due to the strong foundational capabilities of LLMs. We found that directly using LLMs for inference performed worse than the BERT-based methods, but after fine-tuning, the performance drastically improved, indicating that CDA played an indispensable role.

Compared to LLM-based methods, our approach achieved the best performance with the fewest model calls. It is worth mentioning that both LLM-based methods were built upon GPT-3.5, while we achieved better results using only Llama 3.1 8B Instruct. This indicates that our approach does not entirely rely on the foundational capabilities of LLMs; CDA plays a critical role as well.

\subsection{The Performance of CDA on Different LLMs}

\begin{figure}[t]
  \centering        
  \subfloat  
  {
      \includegraphics[width=0.9\linewidth]{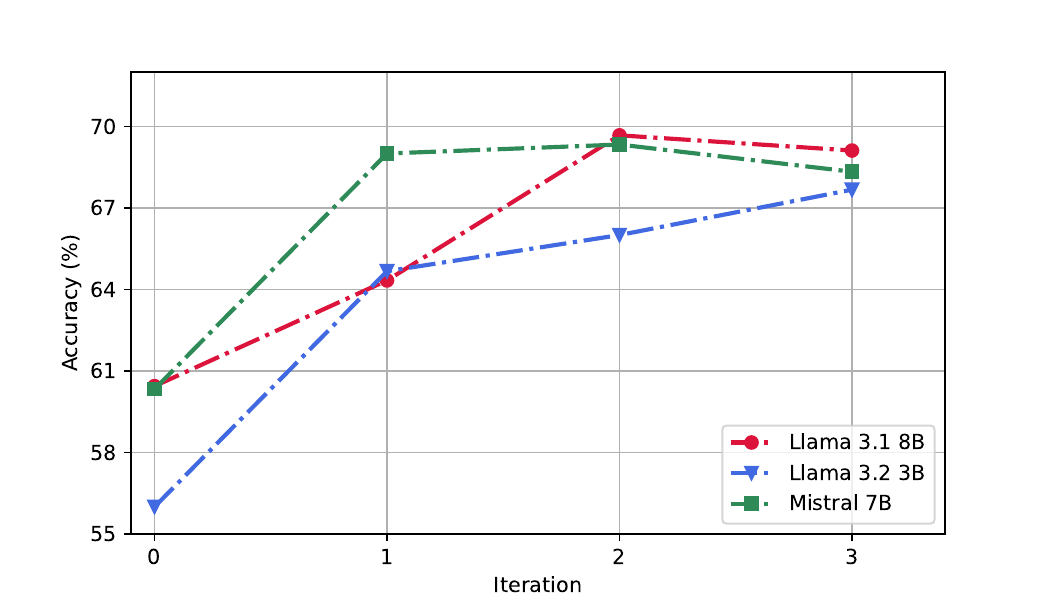}
  }
  \quad
  \subfloat
  {
      \includegraphics[width=0.9\linewidth]{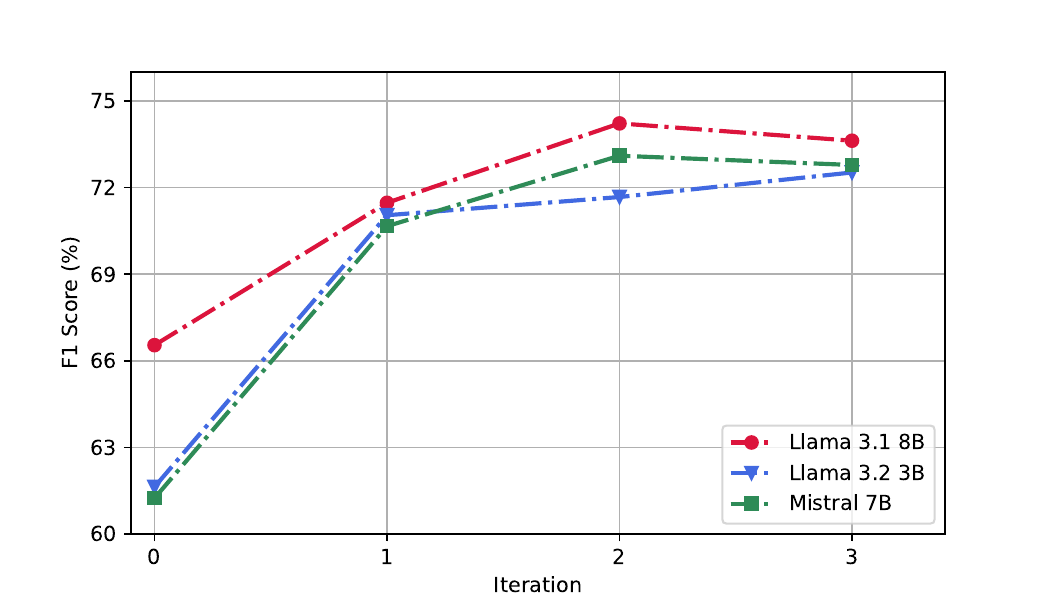}
  }
  \caption{Performance of CDA on different LLMs. This shows the performance of LLMs on the TroFi dataset across different iteration.}    
  \label{llmcmp}            
\end{figure}

\begin{table}
	\centering
	\begin{tabular}{lccc}
		\Xhline{1.2pt}
		\rule{0pt}{15pt}
		\multirow{2}{*}{\textbf{Model}} & \multicolumn{3}{c}{\textbf{\%M}}\\
		\cline{2-4}
		\rule{0pt}{10pt}
		& \uppercase\expandafter{\romannumeral1} & \uppercase\expandafter{\romannumeral2} & \uppercase\expandafter{\romannumeral3} \\
		\hline
		\rule{-3pt}{15pt}
		\textbf{Llama 3.1 8B} & 56.01 & 48.56 & 39.92\\
		\textbf{Llama 3.2 3B} & 56.43  & 50.94 & 41.26\\
		\textbf{Mistral 7B} & 31.56 & 60.63  & 73.32\\
		\Xhline{1.2pt}
	\end{tabular}
	\caption{\label{distribution}
		The data distribution of each model across different iterations, where \textbf{\%M} represents the percentage of metaphorical instances in the training data.}
\end{table}

To demonstrate the effectiveness of CDA, we conducted experiments on Llama 3.1 8B Instruct \citep{llama3}, Mistral 7B Instruct \citep{mistral7b}, and Llama 3.2 3B Instruct \citep{llama3}, and compared the model performance across different iterations. As shown in Figure \ref{llmcmp}, CDA performs well across different LLMs, with a significant improvement in the model's performance.

However, we observed a downward trend in model performance as the number of iterations increased, which is primarily caused by the imbalance in the training data distribution. Although the initial training data distribution is balanced, the distribution of correctly predicted data is not balanced, and this issue becomes more pronounced as the iterations increase. We have summarized the data distribution of the three models across different iterations, as shown in the Table  \ref{distribution}. The data distribution imbalance is more severe in Llama 3.1 8B and Mistral 7B, leading to a performance decline in the final iteration. On the other hand, Llama 3.2 3B exhibits the least data distribution imbalance among the three. Due to its smaller size and relatively weaker learning ability, it is still able to continue learning a substantial amount of knowledge, resulting in improved performance.

\subsection{Ablation Study}

In this section, we provide detailed ablation studies to demonstrate the effectiveness of each component of our method.

\subsubsection{Fine-tuning on the Correct Cases vs. Fine-tuning on All the Cases}

\begin{table}
	\centering
	\begin{tabular}{c|c|cc}
		\Xhline{1.2pt}
		\rule{0pt}{15pt}
		\multirow{2}{*}{\textbf{Dataset}} & \textbf{Only Train} & \multirow{2}{*}{\textbf{Acc.}} & \multirow{2}{*}{\textbf{F1}} \\
		& \textbf{Correct Cases} & & \\
		\hline
		\rule{0pt}{15pt}
		\multirow{2}{*}{\textbf{MOH-X}} & \xmark & 85.33 & 85.53 \\
		& \cellcolor{green!10}{\cmark} & \cellcolor{green!10}{\textbf{86.11}}  &  \cellcolor{green!10}{\textbf{86.57}} \\
		\hline
		\rule{0pt}{15pt}
		\multirow{2}{*}{\textbf{TroFi}} & \xmark & 67.00 &  70.62\\
		& \cellcolor{green!10}{\cmark} & \cellcolor{green!10}{\textbf{69.11}} & \cellcolor{green!10}{\textbf{73.62}} \\
		\Xhline{1.2pt}
	\end{tabular}
	\caption{\label{onlycor}
		Ablation study on whether to only fine-tune on the correct cases. \textbf{Only Train Correct Cases} refers to fine-tuning on the correct cases vs. fine-tuning on all the cases.}
\end{table}

Before each iteration of fine-tuning with CDA, we evaluate the initial data using the student model, and correctly predicted data is selected for fine-tuning. We compared the performance of fine-tuning with the correctly predicted data versus fine-tuning with all the data. As shown in Table \ref{onlycor}, fine-tuning with the correctly predicted data achieved the best results on both datasets, with significantly better performance on the TroFi dataset compared to fine-tuning with all the data. This is because the model, when less capable, encountered more difficult data and failed to learn the challenging knowledge from it, instead only learning relatively simple knowledge. As a result, the model performed poorly on the more difficult dataset.

\subsubsection{Augmenting Only the Wrong Cases vs. Augmenting All the Cases}

\begin{table}
	\centering
	\begin{tabular}{c|c|cc}
		\Xhline{1.2pt}
		\rule{0pt}{15pt}
		\multirow{2}{*}{\textbf{Dataset}} & \textbf{Only Aug.} & \multirow{2}{*}{\textbf{Acc.}} & \multirow{2}{*}{\textbf{F1}} \\
		& \textbf{Wrong Cases} & & \\
		\hline
		\rule{0pt}{15pt}
		\multirow{2}{*}{\textbf{MOH-X}} & \xmark & 86.00 & 86.09 \\
		& \cellcolor{green!10}{\cmark} & \cellcolor{green!10}{\textbf{86.11}}  &  \cellcolor{green!10}{\textbf{86.57}} \\
		\hline
		\rule{0pt}{15pt}
		\multirow{2}{*}{\textbf{TroFi}} & \xmark & 66.33 &  72.02\\
		& \cellcolor{green!10}{\cmark} & \cellcolor{green!10}{\textbf{69.11}} & \cellcolor{green!10}{\textbf{73.62}} \\
		\Xhline{1.2pt}
	\end{tabular}
	\caption{\label{augonly}
		Ablation study on whether to only augment the wrong cases. \textbf{Only Aug. Wrong Cases} refers to augmenting only the wrong cases vs. augmenting all the cases.}
\end{table}

During fine-tuning with CDA, only the data predicted incorrectly by the model is selected for augmentation. We compared the performance of augmenting the incorrectly predicted data versus augmenting all the data, as shown in Table \ref{augonly}. Selecting only the incorrectly predicted data for augmentation achieved the best results on both datasets, with significantly better performance on the TroFi dataset compared to augmenting all the data. This is because selecting incorrect predicted data for augmentation increases the proportion of difficult data, helping the model focus on learning more challenging knowledge. On the other hand, augmenting all the data disperses the model's attention to relatively simpler knowledge, which hinders performance improvement.

\subsubsection{Merged Data as Training Data vs. Augmented Data as Training Data}

\begin{table}
	\centering
	\begin{tabular}{c|c|cc}
		\Xhline{1.2pt}
		\rule{0pt}{15pt}
		\textbf{Dataset} & \textbf{Merge Data} & \textbf{Acc.} & \textbf{F1} \\
		\hline
		\rule{0pt}{15pt}
		\multirow{2}{*}{\textbf{MOH-X}} & \xmark & 78.33 & 75.47 \\
		& \cellcolor{green!10}{\cmark} & \cellcolor{green!10}{\textbf{86.11}}  &  \cellcolor{green!10}{\textbf{86.57}} \\
		\hline
		\rule{0pt}{15pt}
		\multirow{2}{*}{\textbf{TroFi}} & \xmark & \textbf{70.00} &  71.70\\
		& \cellcolor{green!10}{\cmark} & \cellcolor{green!10}{69.11} & \cellcolor{green!10}{\textbf{73.62}} \\
		\Xhline{1.2pt}
	\end{tabular}
	\caption{\label{merge}
		Ablation study on whether to merge the augmented data with the original data as the training data for next iteration. \textbf{Merge Data} refers to selecting the merged data as training data vs. selecting the augmented data as trainging data.}
\end{table}

During fine-tuning with CDA, the augmented data is combined with the original data as the training data for the next iteration. We compared the performance of using merged data as training data versus using only augmented data as training data, as shown in Table \ref{merge}. Although using only the augmented data for training achieved the highest accuracy on the TroFi dataset, it lagged far behind the merged data approach on other metrics. This is because the model focused all its attention on the more difficult knowledge, leading to significant forgetting of simpler knowledge. As a result, the performance on the MOH-X dataset significantly dropped.

\subsubsection{Continuous Fine-tuning vs. From-scratch Fine-tuning}

\begin{table}
	\centering
	\begin{tabular}{c|c|cc}
		\Xhline{1.2pt}
		\rule{0pt}{15pt}
		\multirow{2}{*}{\textbf{Dataset}} & \textbf{Continuous} & \multirow{2}{*}{\textbf{Acc.}} & \multirow{2}{*}{\textbf{F1}} \\
		& \textbf{Fine-tuning} & & \\
		\hline
		\rule{0pt}{15pt}
		\multirow{2}{*}{\textbf{MOH-X}} & \xmark & 85.67 & 85.32 \\
		& \cellcolor{green!10}{\cmark} & \cellcolor{green!10}{\textbf{86.11}}  &  \cellcolor{green!10}{\textbf{86.57}} \\
		\hline
		\rule{0pt}{15pt}
		\multirow{2}{*}{\textbf{TroFi}} & \xmark & 68.33 &  72.78\\
		& \cellcolor{green!10}{\cmark} & \cellcolor{green!10}{\textbf{69.11}} & \cellcolor{green!10}{\textbf{73.62}} \\
		\Xhline{1.2pt}
	\end{tabular}
	\caption{\label{from-scratch}
		Ablation study on whether to fine-tune continuously. \textbf{Continuous Fine-tuning} refers to fine-tuning the model of previous iteration vs. fine-tuning the base model from scratch.}
\end{table}

During fine-tuning with CDA, the model is further fine-tuned based on the checkpoint of previous iteration. We compared the performance of continuous fine-tuning versus fine-tuning from scratch, as shown in Table \ref{from-scratch}. Continuous fine-tuning achieved the best results on both datasets. This is because the model can build on the simple knowledge it has already learned and continue to learn more difficult knowledge, without needing to divert attention to relatively simple data.

\subsubsection{Ablation for CDA}

\begin{table}
	\centering
	\begin{tabular}{c|c|cc}
		\Xhline{1.2pt}
		\rule{0pt}{15pt}
		\textbf{Dataset} & \textbf{Using CDA} & \textbf{Acc.} & \textbf{F1} \\
		\hline
		\rule{0pt}{15pt}
		\multirow{2}{*}{\textbf{MOH-X}} & \xmark & 85.00 & 85.98 \\
		& \cellcolor{green!10}{\cmark} & \cellcolor{green!10}{\textbf{86.11}}  &  \cellcolor{green!10}{\textbf{86.57}} \\
		\hline
		\rule{0pt}{15pt}
		\multirow{2}{*}{\textbf{TroFi}} & \xmark & 65.67 &  70.99\\
		& \cellcolor{green!10}{\cmark} & \cellcolor{green!10}{\textbf{69.11}} & \cellcolor{green!10}{\textbf{73.62}} \\
		\Xhline{1.2pt}
	\end{tabular}
	\caption{\label{lora}
		Ablation study on CDA. \textbf{Using CDA} refers to fine-tuning using CDA vs. fine-tuning using LoRA only.}
\end{table}

In order to explore the overall effect of CDA, we conducted an ablation experiment on CDA. We compared the performance of fine-tuning using CDA versus fine-tuning using LoRA only, as shown in Table       \ref{lora}. On both datasets, the effect of fine-tuning using CDA was much higher than that of fine-tuning using LoRA only, suggesting that our improvement was mainly due to CDA rather than fine-tuning.

\subsubsection{Ablation for Different Augmentation Methods}

\begin{table}
	\centering
	\begin{tabular}{lcccc}
		\Xhline{1.2pt}
		\rule{0pt}{15pt}
		\multirow{2}{*}{\textbf{Aug.}} & \multicolumn{2}{c}{\textbf{MOH-X}} & \multicolumn{2}{c}{\textbf{TroFi}} \\
		\cline{2-5}
		\rule{0pt}{10pt}
		& Acc. & F1 & Acc. & F1\\
		\hline
		\rule{-3pt}{15pt}
		\textbf{w/o Direct} & 86.00 & 86.09 & 68.00 & 72.73\\
		\textbf{w/o Target} & 83.33 & 82.76 & 69.00 & 73.50\\
		\textbf{w/o Context} & 84.33 & 84.28 & 67.67 & 72.21\\
		\hline
		\rule{0pt}{15pt}
		\textbf{Ours} & \textbf{86.11} & \textbf{86.57} & \textbf{69.11} & \textbf{73.62}\\
		\Xhline{1.2pt}
	\end{tabular}
	\caption{\label{augmethods}
		Ablation for different augmentation methods. \textbf{Direct} refers to directly generating a sentence with the target word, \textbf{Target} refers to replacing the target word in the original sentence while keeping its usage unchanged, \textbf{Context} refers to replacing the context in the original sentence while keeping the usage of the target word unchanged, and \textbf{Ours} refers to simultaneously applying all three methods for data augmentation.}
\end{table}

In \ref{dataaug}, we designed three methods for data augmentation specifically for the metaphor detection task. We conducted ablation studies to analyze the effects of these three augmentation methods. As shown in Table \ref{augmethods}, the absence of any one of these augmentation methods leads to a decline in model performance. Specifically, if the method of directly generating sentences with the target word is omitted, there is a significant performance drop on the TroFi dataset, indicating that this augmentation method can generate more challenging data to help the model learn difficult knowledge. If the method of replacing the target word is omitted, there is a significant performance drop on the MOH-X dataset, suggesting that this augmentation method mainly focuses on simpler data and helps the model learn relatively easy knowledge. If the method of replacing the context is omitted, there is a severe performance drop on both datasets, indicating that this augmentation method is the most important, as it helps the model learn both simple and difficult knowledge.

\section{Conclusion}

In this paper, we propose a method for metaphor detection by fine-tuning large language models (LLMs) and introduce \textbf{C}urriculum-style \textbf{D}ata \textbf{A}ugmentation  (\textbf{CDA}), which significantly addresses the data scarcity problem in metaphor detection task. The advantage of this method lies in its ability to quickly learn simple knowledge and gradually acquire more complex knowledge to continuously improve performance. Experimental results demonstrate that our approach achieves state-of-the-art performance across all methods, while also minimizing the number of model calls among LLM-based methods. Furthermore, our method is not limited to metaphor detection and can be generalized to other NLP tasks. Future work can further address the imbalance in data distribution that exists in CDA or explore other ways to extract effective data for data augmentation.

\section*{Limitations}

Our method cannot be iterated indefinitely for data augmentation, which can be attributed to two main reasons: (1) We only select incorrectly predicted data for data augmentation, and the distribution of this data is imbalanced. As the number of iterations increases, the imbalance problem becomes more severe. (2) With more iterations, the number of incorrectly predicted data decreases, so the amount of data generated per iteration gradually decreases.

In addition, we did not conduct experiments on larger models such as Llama 3.1 70B and Llama 3.1 405B, mainly due to computational resource limitations.
\bibliography{anthology,custom}

\clearpage
\appendix

\section{Prompts}
\label{prompts}

In this section, we show the prompts used during data augmentation for metaphor detection.

\begin{table}[H]\footnotesize
\centering
\small
\begin{tabular}{|p{0.95\linewidth}|}
\hline
\rule{-3pt}{12pt}
 You are a creative writing assistant skilled in crafting subtle and intricate metaphors. Your task is to create a sentence that incorporates the metaphorical interpretation of the verb '\{target\_word\}' in an unexpected and unique way. The output must contain the word '\{target\_word\}' as a verb. Please provide only the sentence.\\\\Your sentence:\\
\hline
\end{tabular}
\caption{Prompt for directly generating sentences that include the metaphorical usage of the target word.}
\vspace{-5mm}
\end{table}

\begin{table}[H]\footnotesize
\centering
\small
\begin{tabular}{|p{0.95\linewidth}|}
\hline
\rule{-3pt}{12pt}
 You are a creative writing assistant skilled in crafting subtle and intricate metaphors. Your task is to replace the target word in the following sentence with a new metaphorical expression, ensuring that the new word also carries a metaphorical meaning.\\The following are several examples:\\\\\textbf{Original sentence:} He grasped the concept quickly.\\\textbf{Target word:} grasp\\\textbf{New sentence:} He digested the concept swiftly.\\\textbf{New word:} digest\\\\\textbf{Original sentence:} He soared to new heights in his career.\\\textbf{Target word:} soar\\\textbf{New sentence:} He climbed to new summits in his career.\\\textbf{New word:} climb\\\\\textbf{Original sentence:} \{sentence\}\\\textbf{Target word:} \{target\_word\}\\
\hline
\end{tabular}
\caption{Prompt for retaining the context, replacing the target word, and ensuring that the replaced word still uses a metaphorical meaning.}
\vspace{-5mm}
\end{table}

\begin{table}[H]\footnotesize
\centering
\small
\begin{tabular}{|p{0.95\linewidth}|}
\hline
\rule{-3pt}{12pt}
 You are a creative writing assistant skilled in transforming contexts while preserving metaphorical meanings. Your task is to take the given sentence containing the metaphorical use of the word '\{target\_word\}' and rework it into a new sentence that maintains the metaphorical essence while changing the surrounding context. The output must contain the word '\{target\_word\}'. Please provide only the new sentence.\\\\Given sentence: '\{sentence\}'\\\\Your sentence:\\
\hline
\end{tabular}
\caption{Prompt for retaining the metaphorical usage of the target word and replacing the context.}
\vspace{-5mm}
\end{table}

\begin{table}[H]\footnotesize
\centering
\small
\begin{tabular}{|p{0.95\linewidth}|}
\hline
\rule{-3pt}{12pt}
 You are a straightforward writing assistant skilled in creating clear and literal statements. Your task is to formulate a sentence that uses the verb '\{target\_word\}' in its direct and obvious meaning. The output must contain the word '\{target\_word\}' as a verb. Please provide only the sentence.\\\\Your sentence:\\
\hline
\end{tabular}
\caption{Prompt for directly generating sentences that include the literal usage of the target word.}
\vspace{-5mm}
\end{table}

\begin{table}[H]\footnotesize
\centering
\small
\begin{tabular}{|p{0.95\linewidth}|}
\hline
\rule{-3pt}{12pt}
 You are a straightforward writing assistant skilled in creating clear and literal statements. Your task is to replace the target word in the following sentence with a new literal expression, ensuring that the new word is used in its direct and obvious meaning.\\The following are several examples:\\\\\textbf{Original sentence:} He quickly understood the concept.\\\textbf{Target word:} understand\\\textbf{New sentence:} He quickly comprehended the concept.\\\textbf{New word:} comprehend\\\\\textbf{Original sentence:} She ran fast to catch the bus.\\\textbf{Target word:} run\\\textbf{New sentence:} She sprinted to catch the bus.\\\textbf{New word:} sprint\\\\\textbf{Original sentence:} \{sentence\}\\\textbf{Target word:} \{target\_word\}\\
\hline
\end{tabular}
\caption{Prompt for retaining the context, replacing the target word, and ensuring that the replaced word still uses a literal meaning.}
\vspace{-5mm}
\end{table}

\begin{table}[H]\footnotesize
\centering
\small
\begin{tabular}{|p{0.95\linewidth}|}
\hline
\rule{-3pt}{12pt}
 You are a creative writing assistant skilled in transforming contexts while preserving the literal meanings of words. Your task is to take the given sentence containing the literal use of the word '\{target\_word\}' and rework it into a new sentence that maintains the literal essence while changing the surrounding context. The output must contain the word '\{target\_word\}'. Please provide only the new sentence.\\\\Given sentence: '\{sentence\}'\\\\Your sentence:\\
\hline
\end{tabular}
\caption{Prompt for retaining the literal usage of the target word and replacing the context.}
\vspace{-5mm}
\end{table}

\vspace{5mm}

\section{Statistical Information of The Augmented Data}\label{statistic}

\begin{table}[H]
	\centering
	\resizebox{\linewidth}{!}{
		\begin{tabular}{ccccc}
			\Xhline{1.2pt}
			\rule{0pt}{12pt}
			\textbf{Iteration} & \textbf{\#Instance} & \textbf{\%M} & \textbf{\%Correct} & \textbf{\%Correct.M}\\
			\hline
			\rule{0pt}{12pt}
			\uppercase\expandafter{\romannumeral1} & 6,000 & 50.00 & 68.62  & 56.01 \\
			\uppercase\expandafter{\romannumeral2} & 11,416 & 44.48  & 87.99 & 48.56\\
			\uppercase\expandafter{\romannumeral3} & 15,324 & 36.88 & 88.72 & 39.92 \\
			\Xhline{1.2pt}
	\end{tabular}}
	\caption{\label{augdataset}
		 Statistics of datasets. \textbf{\#Instance}: Number of instances. \textbf{\%M}: Percentage of metaphorical instances. \textbf{\%Correct}: Percentage of correctly predicted cases in all instances. \textbf{\%Correct.M}: Percentage of metaphorical instances in correctly predicted cases.}
\end{table}

\begin{table*}
	\centering
		\begin{tabular}{lcccccc}
			\Xhline{1.2pt}
			\rule{0pt}{15pt}
			\multirow{2}{*}{\textbf{Method}}
			& \multicolumn{2}{c}{\textbf{VUA Adj.}} & \multicolumn{2}{c}{\textbf{VUA Adv.}} & \multicolumn{2}{c}{\textbf{VUA Noun.}}\\ 
			\cline{2-7}
			\rule{0pt}{10pt}
			& Acc. & F1 & Acc. & F1 & Acc. & F1 \\
			\hline
			\rule{-3pt}{15pt}
			Llama 3.1 8B Instruct & 55.84 & 29.31 & 55.47 & 16.01 & 62.74 & 34.21 \\
			Llama 3.1 8B Instruct (CDA) & 76.14 & 42.46 & 85.88 & 35.88 & 80.64 & 52.39\\
			\Xhline{1.2pt} 
	\end{tabular}
	\caption{\label{transfer}
		 Zero-shot experiments conducted on the VUA ALL dataset.}
\end{table*}

Table \ref{augdataset} shows the statistical information of the data across different iterations. Although the distribution of the initial data is balanced, the distribution of correctly predicted data is imbalanced. And as the number of iterations increases, the imbalance in the data distribution becomes more pronounced, which may lead to a decline in model performance. Therefore, for the sake of both data distribution and computational resources, we perform only three iterations.

\section{Zero-shot Transfer on VUA ALL}\label{vuaall}

We extended our experimental scope to other parts of speech within the VUA ALL dataset, encompassing adjectives, adverbs, and nouns. It is important to note that these additional experiments were conducted in a zero-shot manner, meaning that the model was trained exclusively on the verb dataset and then directly tested on datasets of other parts of speech. The experimental results are presented in Table  \ref{transfer}. As can be seen, even when trained solely on the verb dataset, the model's performance on each part-of-speech dataset is significantly enhanced. Thus, our approach is not limited to verb datasets.

\end{document}